\documentclass[letterpaper, 10 pt, journal, twoside]{IEEEtran}  % Comment this line out if you need a4paper

\IEEEoverridecommandlockouts                              % This command is only needed if 
\usepackage{ral}

\newcommand{\isArXiv}{1}

\title{Learning on the Fly: Rapid Policy Adaptation \\via Differentiable Simulation}
\ifthenelse{\equal{\isArXiv}{0}}{
\author{Anonymous Authors}
}
{
\author{Jiahe Pan$^*$, Jiaxu Xing$^*$, Rudolf Reiter, Yifan Zhai, Elie Aljalbout, and Davide Scaramuzza% <-this % stops a space
\thanks{Manuscript received: August 25, 2025; Revised November 14, 2025; Accepted December 17, 2025. This paper was recommended for publication by
Editor Jens Kober upon evaluation of the Associate Editor and Reviewers comments. }% <-this % stops a space
\thanks{
This work was supported by the European Union’s Horizon Europe Research and Innovation Programme under grant agreement No. 101120732 (AUTOASSESS) and the European Research Council (ERC) under grant agreement No. 864042 (AGILEFLIGHT).
*These authors contributed equally to this work.
These authors are with the Robotics and Perception Group, Department of Informatics, University of Zurich,
Switzerland (\protect\url{https://rpg.ifi.uzh.ch}).
        {Contact: \tt\small jixing@ifi.uzh.ch}}%
\thanks{Digital Object Identifier (DOI): see top of this page.}
}
}

\begin{document}

\maketitle
% \thispagestyle{empty}
% \pagestyle{empty}

%===============================================================================

% White paper at RPG:

% 4. Tentative Title (add in comments some alternatives; how to choose a tittle: think of the must-have keywords)

% Tentative abstract:
% 2. Abstract structure (should be less than 200 words):
    % 1. Motivation: Task description / why is it important?
    % 2. Challenge: Why is the problem so difficult?
    % 3. Trends: How does SotA approach it? What's missing?
    % 4. Method: How do you solve it? Contributions!

% 1. Intro structure:
    % 1. What is the problem?
    % 2. Why is it important?
    % 3. Why is the problem hard? What makes it challenging?
    % 4. How far has existing work come? What is the next frontier?
    % 5. Why hasn’t the problem been solved? What is the stumbling block?
    % 6. What does our paper contribute?
    % 7. What is the key idea? What is the magic trick? What is the new insight or technique that enables us to advance the frontier?
    % 8. What do the experiments say?

% 3. Experiments:
    % bullet point description of the planned experiments and justification of each experiment. What is each experiment supposed to validate/prove

%===============================================================================

%%%%%%%%%%%%%%%%%%%%%%%%%%%%%%%%%%%%%%%%%%%%%%%%%%%%%%%%%%%%%%%%%%%%%%%%%%%%%%%%
\vspace{-0.2cm}
\begin{abstract}
Learning control policies in simulation enables rapid, safe, and cost-effective development of advanced robotic capabilities. 
However, transferring these policies to the real world remains difficult due to the sim-to-real gap, where unmodeled dynamics and environmental disturbances can degrade policy performance. 
Existing approaches, such as domain randomization and Real2Sim2Real pipelines, can improve policy robustness, but either struggle under out-of-distribution conditions or require costly offline retraining.
In this work, we approach these problems from a different perspective.
Instead of relying on diverse training conditions before deployment, we focus on rapidly adapting the learned policy in the real world in an online fashion. 
To achieve this, we propose a novel online adaptive learning framework that unifies residual dynamics learning with real-time policy adaptation inside a differentiable simulation.
Starting from a simple dynamics model, our framework continuously refines the model using real-world data to capture unmodeled effects and disturbances, such as payload changes and wind.
The refined dynamics model is embedded in a differentiable simulation framework, enabling gradient backpropagation through the dynamics and thus rapid, sample-efficient policy updates beyond the reach of classical RL methods like PPO.
All components of our system are designed for rapid adaptation, enabling the policy to adjust to unseen disturbances within \emph{5 seconds} of training.
We validate the approach on agile quadrotor control under various disturbances in both simulation and the real world.
Our framework reduces hovering error by up to 81\% compared to $\mathcal{L}_1$-MPC and 55\% compared to DATT, while also demonstrating robustness in vision-based control without explicit state estimation.
\end{abstract}

% Two or three meaningful keywords should be added here
\vspace{-0.2cm}
\hypersetup{
  colorlinks=true,
  urlcolor=ethblue
}

\section*{Supplementary Materials}

\noindent
\faVideo\hspace{0.6em}\text{Video:}  
\url{https://youtu.be/euK2GbcNTvk}

\vspace{0.25em}
\noindent
\faGlobe\hspace{0.6em}\text{Website:}  
\url{https://rpg.ifi.uzh.ch/lotf/}

\vspace{0.25em}
\noindent
\faGithub\hspace{0.6em}\text{Code:}  
\url{https://github.com/uzh-rpg/learning_on_the_fly}
\section{Introduction}\label{sec:introduction}
\IEEEPARstart{R}{obot} learning through simulation has seen great success in recent years, thanks to the rapid improvements in computer hardware and advancements in efficient physics simulation~\cite{collins2021review}.
Simulation provides a fast, safe, and cost-effective way to collect data and train policies, enabling experiments that would be impractical or unsafe in the real world.
However, transferring control policies learned purely in simulation to physical systems is challenging. 
While a high-fidelity simulation model may be used, the system parameters are often difficult to precisely identify. 
In addition, unmodeled effects such as aerodynamic turbulence, sensor noise, and actuator delays further complicate the real-world dynamics, thus making accurate alignment between simulation and reality difficult to achieve.
The resulting mismatch, known as the sim-to-real gap~\cite{aljalbout2025reality}, remains a central obstacle to deploying learning-based controllers in the real world.
Bridging this gap is essential to retain the advantages of simulation while ensuring that policies perform reliably under real-world complexity and variability.

\begin{figure}[t]
    \centering
    \includegraphics[width=1\linewidth]{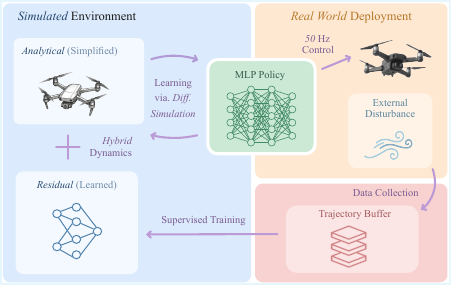}
    \vspace{-0.5cm}
    \caption{Overview of the key components in our proposed approach. (right) The policy is continuously deployed in the real world, and trajectories are collected into a buffer. (bottom) Residual dynamics are trained using the real-world data to refine the \emph{hybrid} simulation dynamics. (left) Based on the latest simulation dynamics, the policy is rapidly adapted via differentiable simulation.}
    \label{fig:overview}
    \vspace{-0.5cm}
\end{figure}

Domain randomization is a common strategy to address this issue~\cite{DR}, in which simulation parameters such as dynamics~\cite{kumar2021rma, kaufmann23champion}, sensor noise~\cite{hwangbo2019learning, openai2019solving}, and visual appearance~\cite{DR, xing2024contrastive} of the environment are varied during training to expose the policy to a wide range of possible deployment scenarios.
By learning in diverse conditions, the agent develops robust policies that are less likely to overfit to the specific characteristics of a single environment.
However, domain randomization cannot exhaustively anticipate all possible real-world conditions.
Thus, when the environment shifts beyond the randomized distribution, policy performance will strongly degrade to out-of-distribution disturbances~\cite{wang2024environment}.
Beyond domain randomization, Real2Sim2Real methods~\cite{hwangbo2019learning, kaufmann23champion} have shown strong sim-to-real transfer capabilities through offline refinement of the simulation model using real-world data before retraining policies for deployment.
While such methods are effective in improving transfer, they require extensive data and costly retraining.
For example,~\cite{bauersfeld2021neurobem} reports 75 minutes of real-world data collection, which makes them unsuitable for rapid online adaptation to changing conditions.

In this work, we approach these problems from another perspective: we propose to \emph{rapidly adapt the policy} to unknown external disturbances in the real world, in an online fashion.
The core insight of the proposed framework is to integrate online residual dynamics learning with rapid policy adaptation via differentiable simulation. 
All system components are designed to update both the residual dynamics and policy \emph{as quickly as possible}, ideally within a few seconds, during runtime.
In this way, the policy becomes adaptive by continuously ``overfitting'' rapidly to the current environment scenario, and paradoxically, more ``generalizable'' across diverse conditions.

For our pipeline, we start with a lightweight rigid-body dynamics model and continuously refine it by learning residual dynamics from real-world flight data.
The residual-augmented dynamics model is embedded in a differentiable simulation framework to achieve more accurate and sample-efficient policy adaptation.
Differentiable simulation provides the key advantage here: by allowing accurate gradients to flow through the dynamics, it makes real-world policy adaptation more efficient than classical RL approaches such as PPO~\cite{song2024learning}.
Another key innovation is our alternating optimization scheme, where policy learning and residual model learning are interleaved so that each batch of real-world data is used efficiently for both dynamics refinement and control improvement through simulation using the refined dynamics.
All of these components ensure the simulation is aligned with reality and enables rapid, efficient policy adaptation to unknown disturbances, even controlling directly from perceptual input. 

We evaluate our proposed framework in both simulation and real-world experiments across various environmental disturbance conditions on an agile quadrotor platform, whose nonlinear dynamics and sensitivity to aerodynamic effects make it a challenging benchmark for adaptive control~\cite{foehn2022agilicious}.
In state-based control tasks such as hovering, where the policy receives as input full quadrotor state information, our method attains an average error of $0.105$\,\si{\meter}, an $81\%$ reduction over $\mathcal{L}_1$-MPC ($0.552$\,\si{\meter}) and $55\%$ over DATT~\cite{huang2023datt} ($0.231$\,\si{\meter}), while ensuring stable flight under modeling errors and out-of-distribution disturbances.
In visual feature-based control, our framework achieves similar gains, demonstrating that rapid adaptation remains effective under partial or noisy observations - a capability unattainable with classical control methods in the absence of state estimation.
\rebuttal{
\subsubsection*{Contributions}
This work aims to enable rapid, real-time policy adaptation on real robotic systems by integrating differentiable simulation and online residual dynamics learning. 
To the best of our knowledge, this is the first demonstration of coupling a differentiable simulator with real-time residual learning to achieve rapid, closed-loop policy adaptation in the real world.
The framework continuously calibrates a lightweight analytical dynamics model with real-world data, allowing fast and stable adaptation to unseen disturbances. 
An alternating optimization scheme interleaves residual model updates and policy learning in a closed-loop manner, ensuring each real-world sample contributes to both model fidelity and control performance. 
Several design choices, including backpropagation through the analytical model only and an asynchronous implementation for concurrent data collection and optimization, were critical to achieving this speed and robustness.
The framework supports both state-based and visual feature-based inputs (without explicit state estimation), and we demonstrate its effectiveness through simulated and real-world experiments, where it outperforms both classical and learning-based controllers under large unseen disturbances.
Together, our framework demonstrates that policies can learn and adapt within seconds in the real world, thereby reducing reliance on domain randomization, which can fail to capture real-world complexity and adapt to out-of-distribution scenarios.
}

\begin{figure}[t]
    \centering
    \includegraphics[width=\columnwidth]{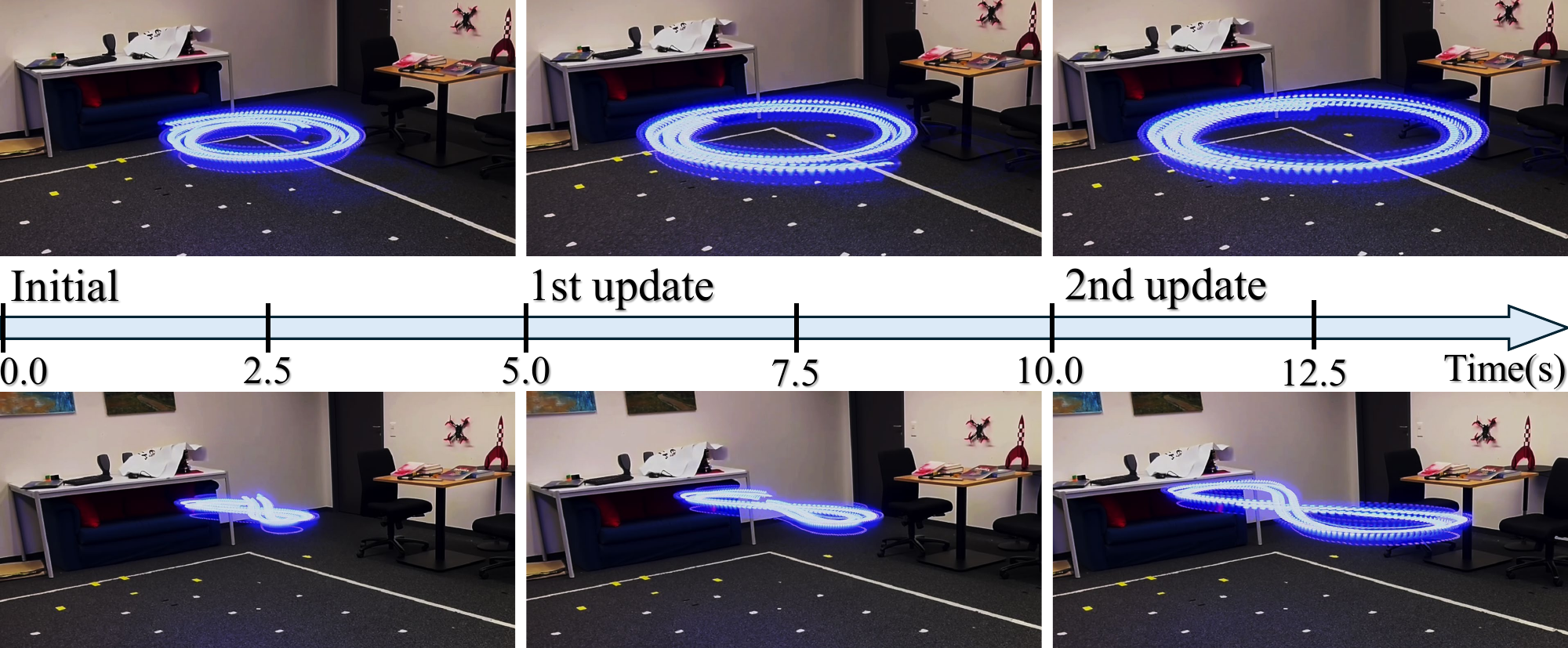}
    \vspace{-0.4cm}
    \caption{Real-world trajectory tracking adaptation using our proposed approach. The policy rapidly learns within 2 updates (10\,\si{\second} of flight) to compensate for the large sim-to-real gap caused by model mismatch in pretraining (see Sec.\,\ref{sec:methodology}).}
    \label{fig:real_world_tracking}
    \vspace{-0.6cm}
\end{figure}
\vspace{-0.2cm}
\section{Related Works}\label{sec:related_work}
\subsection{Aligning Simulation with the Real-World} 
Closing the sim-to-real gap requires quantifying the misalignment between simulation and real-world dynamics, typically through system identification or residual dynamics learning. 
System identification estimates parameters of an analytical dynamics model from input–output data~\cite{ljung1999system}, but its representation capacity is limited to the modeled system~\cite{heiden2021neuralsim}, making it less effective for capturing complex dynamics and disturbances. 
Residual dynamics learning addresses this by directly modeling the discrepancy between analytical predictions and real-world measurements. 
It has been applied to improve quadrotor odometry and tracking~\cite{bauersfeld2021neurobem, torrente2021data}, learn motor delays in quadrupeds~\cite{hwangbo2019learning}, and predict residual forces in soft robots~\cite{gao2024sim}.
\vspace{-0.4cm}
\subsection{Fast Policy Learning in Simulation}
While traditional RL methods suffer from prohibitively long training times, significant speed-ups can be achieved using highly optimized physics simulation~\cite{eschmann2024learning}. 
However, these methods are limited by sample inefficiency due to the high variance in the zeroth-order policy gradient estimates.
A recent alternative paradigm, policy learning via differentiable simulation, uses smooth, differentiable dynamics and rewards to enable policy learning via \emph{first-order} gradients~\cite{newbury2024review}, offering substantial gains in sample efficiency and training time over RL~\cite{heeg2024learning}. 
It has been applied to direct policy parameterizations, such as parametric curve frequencies for swimming robots~\cite{nava2022fast} and sinusoidal policies for robotic cutting~\cite{heiden2021disect}. %
For applications to neural network policies, however, unstable gradients often limit applications to short-horizon tasks with simplified contacts and restricted start-state variation~\cite{xu2023efficient}.
To address this, prior work has explored enhancements such as early-stopping simulations at contact events, truncated BPTT~\cite{xu2022accelerated}, and reward augmentation with a learned critic~\cite{georgiev2024adaptive, luo2024residual}.
\vspace{-0.4cm}
\subsection{Learning-Based Adaptive Control}
Residual dynamics models have been used for online disturbance estimation via offline-trained networks~\cite{shi2019neural}, Gaussian Processes~\cite{torrente2021data}, or differentiable simulation–based system identification~\cite{chen2022real}. 
However, these methods mainly augment optimization-based controllers like MPC, which rely on full state information, and do not directly extend to vision-based control. 
Neural network policies have also been conditioned on disturbance estimates~\cite{huang2023datt,o2022neural,kumar2021rma}, but since they are trained offline in randomized simulations and remain fixed during deployment, they struggle with domain shifts and unseen real-world conditions~\cite{xing2024multitask, huang2023datt}.
\vspace{-0.7em}

\section{Methodology}\label{sec:methodology}

\begin{figure*}[ht]
    \centering
    \includegraphics[width=0.96\linewidth]{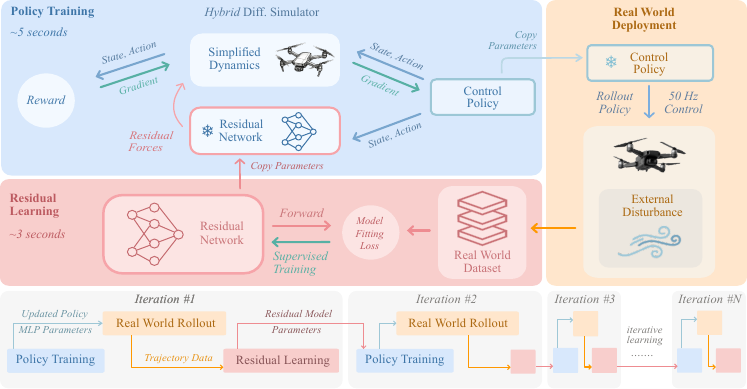}
    \caption{\rebuttal{Detailed illustration of the information flow within and between the three interleaved components of the proposed framework, including residual dynamics learning, differentiable simulation, and policy adaptation. These components operate concurrently across multiple threads in separate ROS nodes.}}
    \label{fig:overall_online_framework}
    \vspace{-0.6cm}
\end{figure*}

Our approach consists of two phases: policy pretraining and online adaptation.
During pretraining, we train a base policy for online adaptation using a low-fidelity analytical dynamics model \emph{without} residual dynamics.
During online adaptation (see Fig.\,\ref{fig:overall_online_framework}), residual dynamics learning, policy adaptation, and real-world deployment run in parallel across multiple threads, with parameters exchanged efficiently between processes via ROS as serialized byte strings.
The residual dynamics network is continuously updated from a rolling buffer of quadrotor states and actions, and combined with the analytical model to form a hybrid dynamics embedded in the differentiable simulation for policy adaptation.
The deployment loop uses the latest policy network parameters to continuously output control commands at 50\,\si{\hertz} to the on-board flight controller, and simultaneously collects flight trajectories.
\vspace{-0.4cm}

%%%%%%%%%%%%%%%%%%%%%%%%%%%%%%%%%%%%%%%%%%%%%%%%%%%%%%%%%%%%%%%%%%
\subsection{Differentiable Simulation Model}
We model the quadrotor as a discrete-time dynamical system with continuous state and action spaces $\mathcal{X}$ and $\mathcal{U}$, respectively. 
The system evolves according to the differentiable hybrid dynamics model $f_{\text{hybrid}}: \mathcal{X} \times \mathcal{U} \mapsto \mathcal{X}$ which comprises the analytical and learned residual components, and describes the system evolution $x_{t+1}=f_{\text{hybrid}}(x_t, u_t)$ over time. 
At time step $t$, an observation model $h: \mathcal{X} \mapsto \mathcal{O}$ generates an observation $o_t=h(x_t)\in\mathcal{O}$ from the state $x_t$, and is passed as input to a deterministic and differentiable policy network $\pi_\phi: \mathcal{X} \mapsto \mathcal{U}$ which outputs an action $u_t = \pi_\phi(o_t)$, and finally a deterministic, smooth and differentiable reward function $r: \mathcal{X} \times \mathcal{U} \mapsto \mathbb{R}$ emits a reward $r_t = r(x_t, u_t)$ based on the state-action pair. 
Thus, all components are fully differentiable and allow gradient backpropagation through the simulation.
\vspace{-0.4cm}

\subsection{Low-Fidelity Quadrotor Dynamics Model}
\label{subsec:lowfid_quad_dynamics}
Given the quadrotor state $\bm{x}$ consisting of position $\bm{p} \in \mathbb{R}^3$, rotation matrix $\bm{R} \in \text{SO}(3)$, and linear velocity $\bm{v} \in \mathbb{R}^3$, and commands $\bm{u}$ consisting of the mass-normalized collective thrust $c \in \mathbb{R}$ and body rates $\bm{\omega}_{\text{cmd}}\in\mathbb{R}^3$, the low-fidelity, analytical quadrotor dynamics $f_{\text{a}}$ is defined as
\begin{equation}
\begin{aligned}
    \dot{\bm{x}} = \frac{d}{dt} \begin{bmatrix} \bm{p}\\ \text{vec}(\bm{R})\\ \bm{v} \end{bmatrix} = \begin{bmatrix} \bm{v}\\ \text{vec}(\bm{R} [\bm{\omega_{\text{cmd}}}]_\times)\\ \bm{R}\bm{c}+\bm{g} \end{bmatrix} \coloneq f_{\text{a}}(\bm{x}, \bm{u}),
\end{aligned}
\label{eqn:simple_quad_dynamics}
\end{equation}
where $[\cdot]_\times$ denotes the skew-symmetric matrix operator and $\text{vec}(\cdot)$ indicates vectorization of a matrix, $\bm{c} = [0, 0, c]^\top$ is the collective thrust vector, and $\bm{g}$ is the gravity vector.
\vspace{-0.3cm}

%%%%%%%%%%%%%%%%%%%%%%%%%%%%%%%%%%%%%%%%%%%%%%%%%%%%%%%%%%%%%%%%%%
\subsection{Policy Optimization Using Analytical Gradients}
The policy learning objective is to maximize the cumulative task reward $\mathcal{R}(\phi)$ over an $N$-step rollout of the policy parameterized by $\phi$ via
\begin{equation}
    \max_\phi \mathcal{R}(\phi) = \sum_{t=0}^{N-1}r(x_t, u_t) = \sum_{t=0}^{N-1}r(x_t, \pi_\phi(h(x_t))).
\label{eqn:reward_objective}
\end{equation}
By leveraging the differentiable dynamics and reward structure, we can obtain first-order analytical policy gradients of the objective \eqref{eqn:reward_objective} via Back-Propagation Through Time (BPTT) (see \cite{metz2021gradients} for a full derivation).
The gradient and the update rule of the policy parameters $\phi$ are given by
\begin{equation}
\begin{aligned}
    \nabla_\phi R(\phi) &= \frac{1}{N}\sum_{t=0}^{N-1} ( \sum_{i=1}^t \frac{\partial r_t}{\partial x_t} \prod_{j=1}^t (\frac{dx_j}{dx_{j-1}}) \frac{\partial x_i}{\partial \phi} + \frac{\partial r_t}{\partial u_t} \frac{\partial u_t}{\partial \phi} ), \\
    \phi_{k+1} &= \phi_k + \alpha \nabla_\phi R(\phi_k),
\end{aligned}
\label{eqn:policy_gradient}
\end{equation}
where $\frac{dx_j}{dx_{j-1}}$ is the derivative matrix of the system dynamics $f_{\text{hybrid}}$, and $\alpha$ is the learning rate. 
We build upon an existing open-source differentiable simulator for quadrotors~\cite{heeg2024learning} written entirely in JAX to leverage both its automatic-differentiation framework for computing the analytical policy gradients and performing GPU-accelerated parallel simulation.
\vspace{-0.5cm}
%

%%%%%%%%%%%%%%%%%%%%%%%%%%%%%%%%%%%%%%%%%%%%%%%%%%%%%%%%%%%%%%%%%%
\subsection{Residual Dynamics Learning}
Given the concatenated input vector $[\bm{x}^\top, \bm{u}^\top] \in \mathbb{R}^{19}$ of quadrotor state $\bm{x}^\top = [\bm{p}^\top, \text{vec}({\bm{R}})^\top, \bm{v}^\top] \in \mathbb{R}^{15}$ and action $\bm{u}^\top = [c, \bm{\omega}_{\text{cmd}}^\top] \in \mathbb{R}^4$, 
an MLP network $f_{\text{res}}$ parameterized by $\bm{\theta}$ is trained to predict the residual acceleration $\bm{a}_{\text{res}} \in \mathbb{R}^3$, defined as the difference between the ground-truth acceleration $\bm{a}_{\text{gt}} \in \mathbb{R}^3$ measured on the real system and the theoretical acceleration $\hat{\bm{a}} \in \mathbb{R}^3$ from the analytical dynamics $f_{\text{a}}(\bm{x}, \bm{u})$ in~\eqref{eqn:simple_quad_dynamics}.
The residual acceleration training targets are computed as $\bm{a}_{\text{res}} = \bm{a}_{\text{gt}} - \hat{\bm{a}}$. 
Given a batch of $|\mathcal{B}|$ samples $\{[\bm{x}^\top, \bm{u}^\top]^i, \bm{a}_{\text{res}}^i\}_{i\in \mathcal{B}}$, we train the model by minimizing the loss function $\mathcal{L}_{\text{res}}$ via
$
    \min_\theta \mathcal{L}_{\text{res}} = \min_\theta \frac{1}{|\mathcal{B}|}\sum^{|\mathcal{B}|}_{i=1}|| \bm{a}_{\text{res}}^i - f_{\text{res}}([\bm{x}^\top, \bm{u}^\top]^i; \bm{\theta}) ||^2 + \beta\sum^L_{l=1} || W^l ||^2_2,
$
where $W^l$ is the weight matrix of the $l$-th network layer, and $\beta$ controls the regularization strength.
The loss comprises a standard MSE term and a spectral norm regularization term, where the latter has been shown to improve generalization beyond the training distribution~\cite{bartlett2017spectrally} by regulating the network's Lipschitz constant~\cite{shi2019neural}. 
\vspace{-0.3cm}

%%%%%%%%%%%%%%%%%%%%%%%%%%%%%%%%%%%%%%%%%%%%%%%%%%%%%%%%%%%%%%%%%%
\subsection{Design Choices for Maximum Runtime Efficiency}
During forward simulation, we use a hybrid dynamics model $f_{\text{hybrid}}$ obtained by additively combining the analytical $f_{\text{a}}$ and learned residual $f_{\text{res}}$ dynamics models. 
Here, we use a simple, low-fidelity analytical dynamics model (see Sec.\,\ref{subsec:lowfid_quad_dynamics}) which models the quadrotor as a point-mass. 
The resulting acceleration given a state and action input pair is computed as $\hat{\bm{a}}_{\text{hybrid}} = \hat{\bm{a}} + \hat{\bm{a}}_{\text{res}}$, where $\hat{\bm{a}}_{\text{res}}$ is the network prediction.
The quadrotor states are simulated at 50\,\si{\hertz} via Runge-Kutta 4 time-integration of the dynamics using $\hat{\bm{a}}_{\text{hybrid}}$. 
While the hybrid dynamics model $f_{\text{hybrid}}$ composed of differentiable analytical and learned components remains overall fully differentiable, we only perform gradient backpropagation through the analytical dynamics model and not the frozen network to obtain the policy gradients. 
This was inspired by prior work in policy learning using differentiable simulation for both quadruped~\cite{song2024learning} and quadrotor~\cite{heeg2024learning} control, which showed that combining accurate forward dynamics simulation with the backpropagation of a surrogate gradient based on a simplified dynamics model achieves faster runtime without impacting the resulting policy performance. We analyze and justify the above design choices through simulated experiments, and present and discuss the results in Sec.\,\ref{subsec:design_choice_justification}. 
\vspace{-0.3cm}
%%%%%%%%%%%%%%%%%%%%%%%%%%%%%%%%%%%%%%%%%%%%%%%%%%%%%%%%%%%%%%%%%%
\subsection{Full vs. Low-Rank Policy Adaptation} 
We compare two existing methods of adapting a pretrained policy: full vs. low-rank adaptation (LoRA)~\cite{bo20243d}.
Full adaptation involves updating all parameters of the policy network, similar to~\cite{xing2024bootstrapping}, whereas LoRA freezes all pretrained parameters and instead adapts an additive low-rank network module~\cite{han2024parameter} which forms a lower-dimensional trainable parameter space.
The latter has been shown to achieve more memory and parameter-efficient policy adaptation using RL for task-transfer~\cite{bo20243d} and multi-agent~\cite{zhang2025low} learning.
Therefore, as an exploratory comparison, we seek to understand whether LoRA can also be combined with sample-efficient policy learning using differentiable simulation to effectively adapt a pretrained policy to unknown environmental disturbances.
\vspace{-0.3cm}

\section{Experiments}\label{sec:experiments}

%%%%%%%%%%%%%%%%%%%%%%%%%%%%%%%%%%%%%%%%%%%%%%%%%%%%%%%%%%%%%%%%%%
\subsection{Experimental Setup}
\begin{figure}[b]
    \centering
    \hfill
    \begin{subfigure}{0.3\columnwidth}
        \centering
        \includegraphics[width=\linewidth]{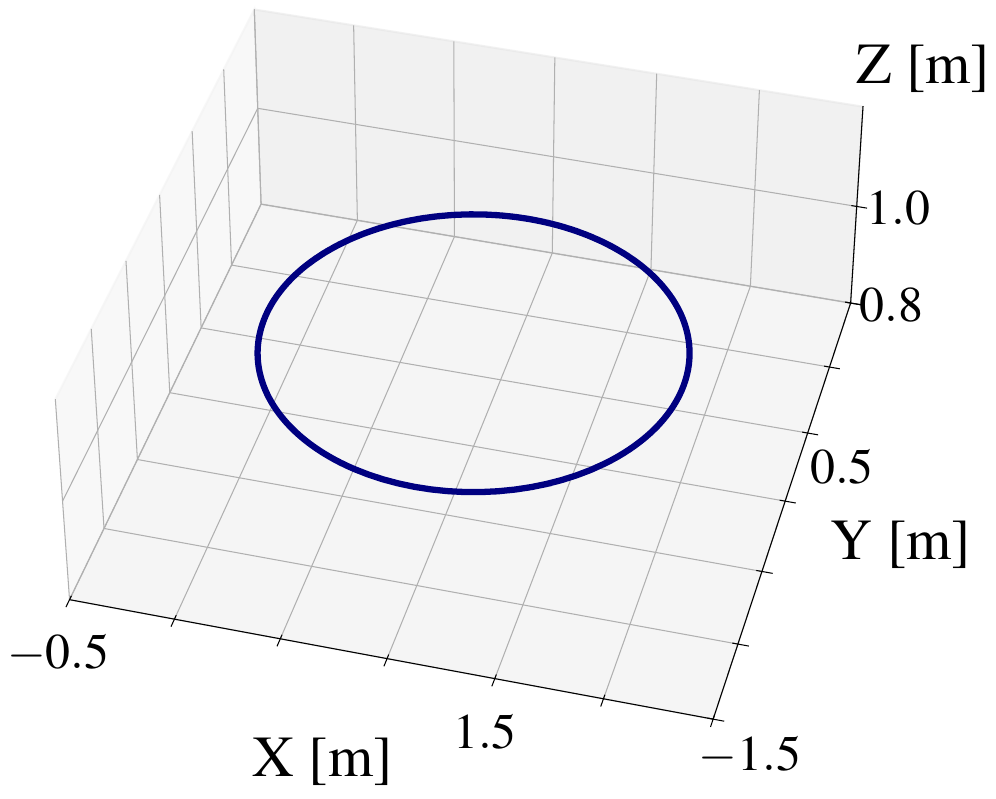}
        \vspace{-0.5cm}
        \caption{\footnotesize \textit{Circle}}
        \label{fig:circle_ref_traj}
    \end{subfigure}
    \hfill
    \begin{subfigure}{0.3\columnwidth}
        \centering
        \includegraphics[width=\linewidth]{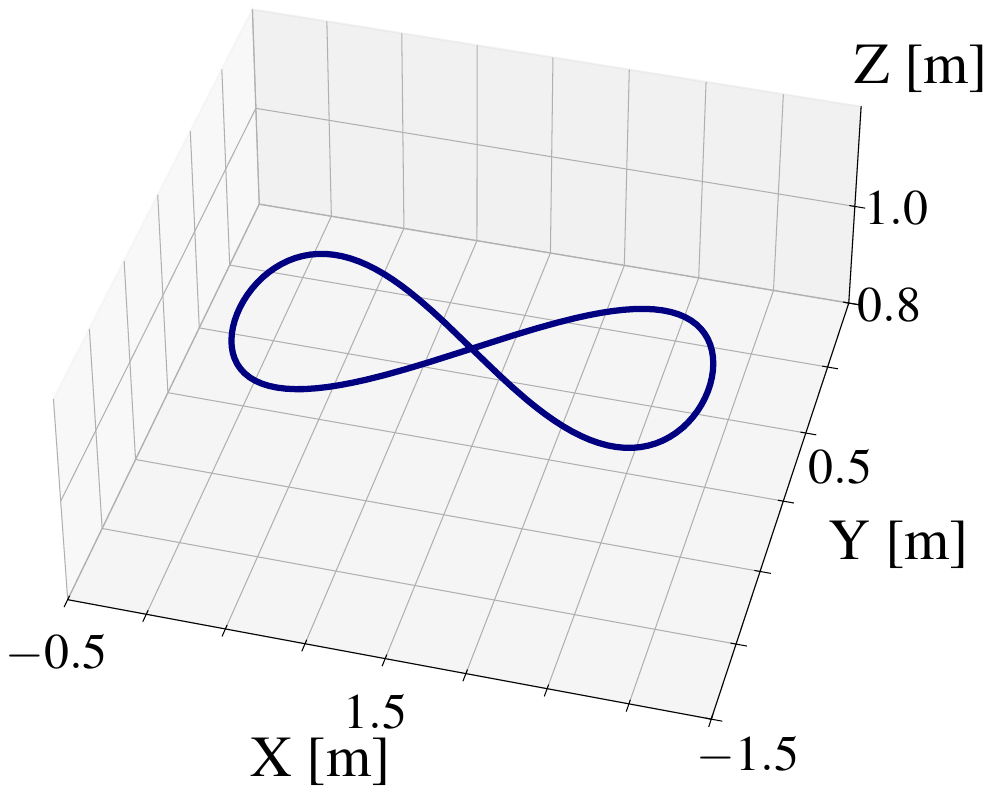}
        \vspace{-0.5cm}
        \caption{\footnotesize \textit{Figure-8}}
        \label{fig:fig8_ref_traj}
    \end{subfigure}
    \hfill
    \begin{subfigure}{0.3\columnwidth}
        \centering
        \includegraphics[width=\linewidth]{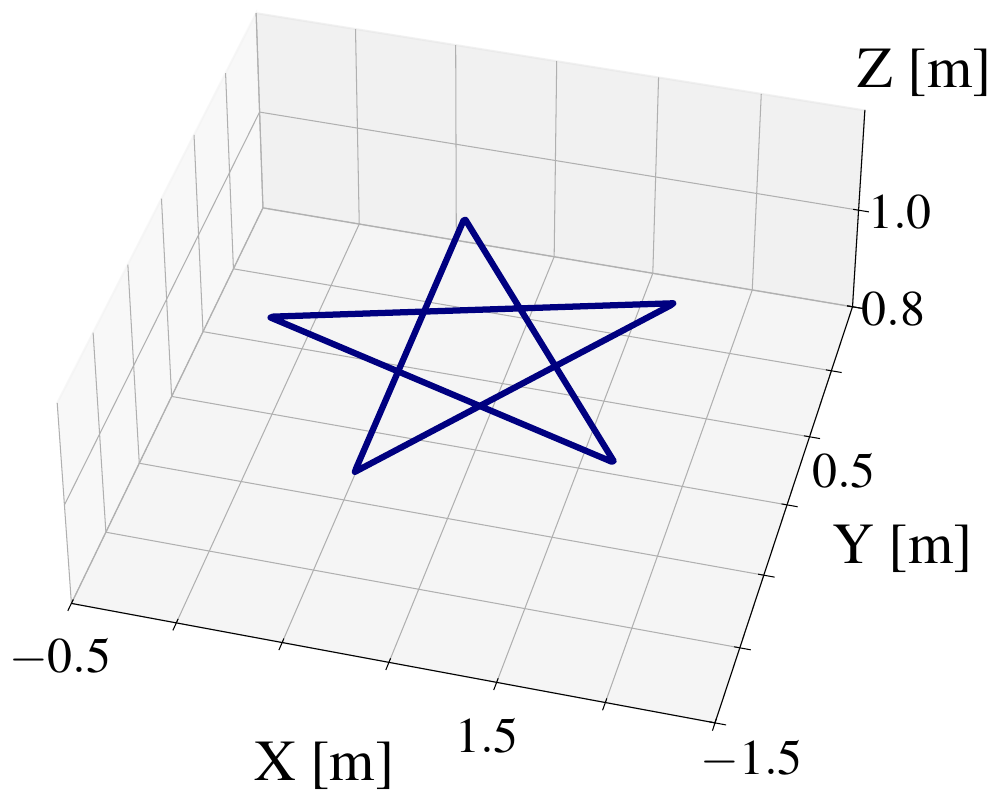}
        \vspace{-0.5cm}
        \caption{\footnotesize \rebuttal{\textit{5-Point Star}}}
        \label{fig:star_ref_traj}
    \end{subfigure}
    \hfill
    \vspace{-0.1cm}
    \caption{The \textit{Circle}, \textit{Figure-8} and \rebuttal{\textit{5-Point Star}} reference trajectories, with periods of 3\,\si{\second}, 5\,\si{\second} and \rebuttal{6\,\si{\second}} respectively. All trajectories lie in the horizontal xy-plane 1\,\si{\meter} above the ground, and start at the point $(0, 0, 1)$\,\si{\meter}.}
    \label{fig:ref_trajs}
    \vspace{-0.4cm}
\end{figure}
\subsubsection{Task and Reward Definitions} 
We evaluate our approach on quadrotor stabilizing hover and trajectory tracking. 
For stabilizing hover, the policy is required to regulate the quadrotor towards a goal $\bm{p}_{\rm{des}}$ and maintain it at all times, which is non-trivial given the quadrotor's non-linear and unstable dynamics. 
We evaluate both a state-based policy which receives observations $\bm{o} = [\bm{p}, \bm{R}, \bm{v}]^\top$ at each time step, and an end-to-end visual feature-based policy which only receives the projected pixel coordinates of seven 3D keypoints from the past five time steps and the last three actions. 
For real-world experiments, the 3D keypoints are simulated in a hardware-in-the-loop style using quadrotor state estimates from a motion-capture system.
Our training setup closely follows the open-source environment setup in~\cite{heeg2024learning}.
Trajectory tracking requires following a reference trajectory defined as a time-parameterized sequence of quadrotor states, with policy training done in a state-based setting. 
As shown in Fig.\,\ref{fig:ref_trajs}, we generate two smooth trajectories, \textit{Circle} and \textit{Figure-8}, \rebuttal{and a non-smooth \textit{5-Point Star} trajectory featuring a highly discontinuous velocity profile.}
For both tasks, the reward at each time step $t$ is defined as a sum of position, velocity, and actuation rewards $
    r_t = r_t^{\rm{pos}} + r_t^{\rm{vel}} + r_t^{\rm{act}}.
$
For stabilizing hover, the individual reward terms are
$    r_t^{\rm{pos}} = -1.0 \cdot L_{\rm{H}}(5 \cdot (\bm{p}_t - \bm{p}_{\rm{des}})), 
    r_t^{\rm{vel}} = -0.1 \cdot L_{\rm{H}}(\bm{v}_t) - 0.1 \cdot L_{\rm{H}}(\bm{\omega}_t)$, and $ 
    r_t^{\rm{act}} = -0.5 \cdot L_{\rm{H}}(\bm{u}_t - \bm{u}_{\rm{hover}}),$
where $L_{\rm{H}}$ is the Huber loss, and $\bm{u}_{\rm{hover}} = [9.81, 0, 0, 0]^\top$ is the mass-normalized action required to counteract gravity. 
For trajectory tracking, the individual reward terms are
   $ r_t^{\rm{pos}} = -1.0 \cdot L_{\rm{H}}(\bm{p}_t - \bm{p}_t^{\rm{ref}}), 
    r_t^{\rm{vel}} = -1.0 \cdot L_{\rm{H}}(\bm{v}_t - \bm{v}_t^{\rm{ref}}),$ and $
    r_t^{\rm{act}} = -0.1 \cdot L_{\rm{H}}(\bm{u}_t - \bm{u}_{\rm{hover}}),$
where $\bm{p}_t^{\rm{ref}}$ and $\bm{v}_t^{\rm{ref}}$ are respectively the reference position and velocity at time $t$. 
\subsubsection{Pretraining Phase} 
We parameterize the policy as an MLP with two 512-dim hidden layers. 
For both state-based hovering and trajectory tracking, we train the base policy from random initialization for 300 epochs across 100 parallel environments.
Each epoch lasts 3 seconds or 150 simulation steps. 
For visual feature-based hovering, we use the initialization approach from~\cite{heeg2024learning} to first train a neural network on a state-representation learning task and use the learned parameters to partially initialize the policy network. 
We refer the reader to~\cite{heeg2024learning} for details on this initialization method.
We then train the partially initialized policy for 500 epochs across 300 parallel environments.

\subsubsection{Online Adaptation Phase} 
The quadrotor states and actions are continuously recorded into a rolling history buffer at 50\,\si{\hertz} and are used to train the residual dynamics network. 
For stabilizing hover and trajectory tracking, we use history buffer sizes of 100 and 250, equivalent to 2 and 5 seconds of trajectory history, respectively. 
For residual dynamics learning, we continuously refine an ensemble of 3 networks, each with two 128-dim hidden layers and initialized using different random seeds, and use the empirical mean prediction from all models as the final predicted residual acceleration for a given input.
Empirically, we found this to effectively reduce the prediction variance arising from epistemic uncertainty due to the limited samples in the data buffer. 
\rebuttal{For LoRA, we follow the original implementation and initialization in~\cite{hu2022lora}, and use ranks of 4, 10, and 1 for the three weight matrix layers of the policy network, respectively, and a constant scale of 4 across all layers.}
We run residual dynamics learning every 3 seconds and train the ensemble networks in parallel for 100 iterations.
Policy adaptation is run every 5 seconds, and we train the state-based policy with 10 parallel simulated environments for 30 epochs and the vision-based policy with 30 environments for 50 epochs. These values were empirically found to provide a good balance between training time and policy performance.
\vspace{-0.3cm}
\begin{table}[t]
\centering
\caption{Average steady-state error (in \si{\meter}) from the hovering target across 8 rollouts. The errors of the two best-performing methods for each disturbance condition are highlighted in green and orange.}
\label{table:state_hovering_results}
\setlength{\tabcolsep}{4pt}
\vspace{-0.1cm}
\begin{tabular}{cccc}
\toprule
Method        & No Dist. & Small Dist. & Large Dist. \\ \midrule
Base DiffSim          & 0.128 $\pm$ 0.004   & 0.328 $\pm$ 0.001      & 1.228 $\pm$ 0.073      \\
$\mathcal{L}_1$-MPC        & 0.091 $\pm$ 0.052   & 0.134 $\pm$ 0.073      & 0.552 $\pm$ 0.130       \\
DATT (PPO)          & \cellcolor{YellowGreen!25} {0.013} $\pm$ 0.004   & \cellcolor{Orange!10} 0.009 $\pm$ 0.005      & 0.231 $\pm$ 0.004      \\
Ours          & \cellcolor{Orange!10} 0.015 $\pm$ 0.001   & \cellcolor{YellowGreen!25} 0.008 $\pm$ 0.002      & \cellcolor{YellowGreen!25} 0.105 $\pm$ 0.007      \\
Ours (LoRA) & 0.023 $\pm$ 0.002   & 0.015 $\pm$ 0.004      &\cellcolor{Orange!10} 0.125 $\pm$ 0.002      \\ \bottomrule
\end{tabular}
\vspace{-0.6cm}
\end{table}

%%%%%%%%%%%%%%%%%%%%%%%%%%%%%%%%%%%%%%%%%%%%%%%%%%%%%%%%%%%%%%%%%%
\subsubsection{Baselines Methods} 
We compare against a state-of-the-art learning-based adaptive control method, Deep Adaptive Tracking Control (DATT)~\cite{huang2023datt}, which uses the popular model-free RL algorithm PPO with domain randomization and online $\mathcal{L}_1$ adaptive control-based disturbance estimation.
For quadrotor control, this method has been shown to outperform Rapid Motor Adaptation (RMA)~\cite{kumar2021rma}, which is a similar approach but instead uses a learned encoder for disturbance estimation.
We used the open-source implementation of~\cite{huang2023datt} and the exact same training procedure and hyperparameters to train both state-based hovering and trajectory tracking policies using PPO for 20 million simulation steps.
Using their original domain randomization method, we simulated 3-dimensional acceleration disturbances as random walks within the bounds +/- $[1, 1, 1] \, \unit{m/s^2}$.
We also compare against an adaptive Nonlinear MPC controller ($\mathcal{L}_1$-MPC) as implemented in~\cite{huang2023datt}, which uses a Model Predictive Path Integral (MPPI) formulation and the same $\mathcal{L}_1$ adaptive control-based disturbance estimation as in DATT.
Additionally, we include our pretrained base policy (Base DiffSim) without online adaptation for comparison.
\vspace{-0.3cm}
%
%%%%%%%%%%%%%%%%%%%%%%%%%%%%%%%%%%%%%%%%%%%%%%%%%%%%%%%%%%%%%%%%%%
\subsection{Experimental Results}
We used a realistic quadrotor simulator~\cite{foehn2022agilicious} equipped with the BEM model for aerodynamic effects and high-frequency simulation of controller dynamics. 
We simulated three levels of constant, uniform acceleration disturbances: $[0, 0, 0]\,\unit{m/s^2}$\,(\textit{none}), $[0.5, 0.5, 0.5]\,\unit{m/s^2}$\,(\textit{small}), and $[2, 2, 2] \,\unit{m/s^2}$\,(\textit{large}).
The first two conditions are within the domain randomization range used for DATT training, whereas the third condition was deliberately chosen to be out of distribution to evaluate its generalization capabilities.
All experiments (simulated and real-world) were run using an Nvidia RTX\,4090 GPU (24\,GB VRAM) with an Intel 14900KF CPU. 

\subsubsection{Performance Comparison to Baseline Approaches}
\label{subsubsec:baseline_comparisons}
\begin{figure*}[ht]
    \centering
    \hspace{-0.1cm}
    \hfill
    \begin{subfigure}{0.95\columnwidth}
        % \centering
        \includegraphics[width=\linewidth]{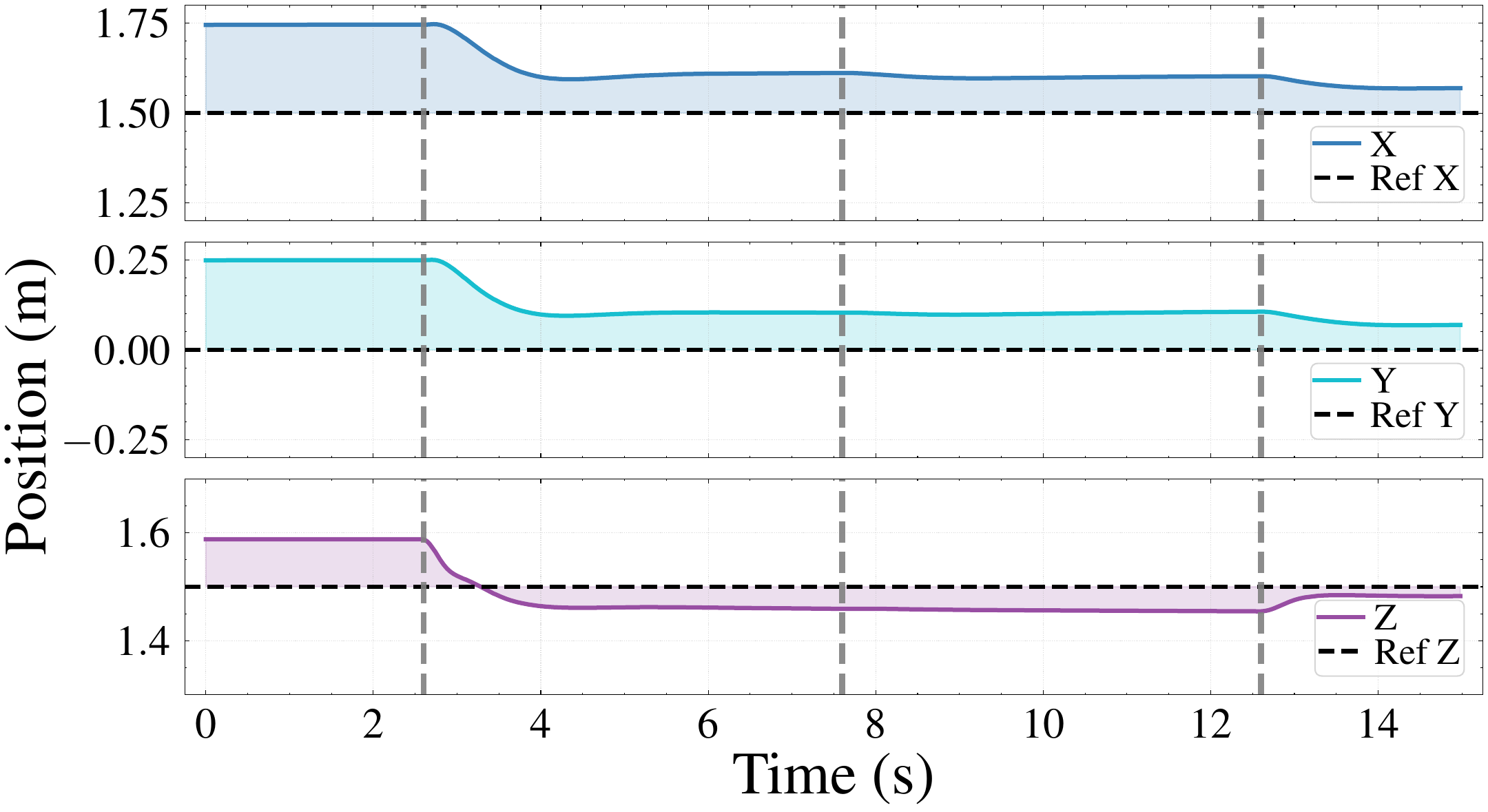}
        \caption{\footnotesize State-based hovering adaptation to constant \textit{large} disturbance.}
        \label{fig:sim_hovering}
    \end{subfigure}
    \hspace{0.1cm}
    \hfill
    \hspace{0.1cm}
    \begin{subfigure}{0.9\columnwidth}
        % \centering
        \includegraphics[width=\linewidth]{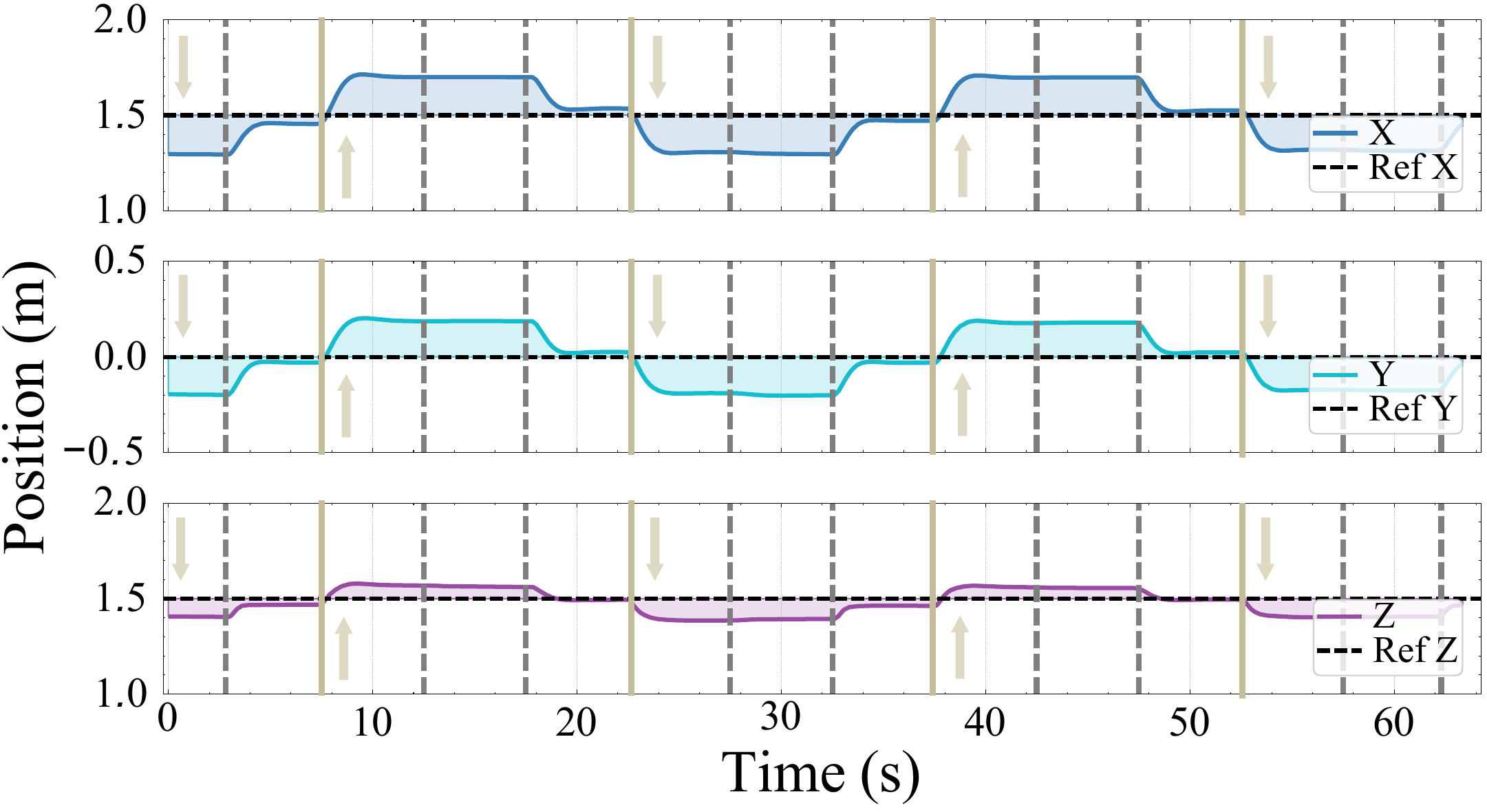}
        \caption{\footnotesize Continuous state-based hovering adaptation to \emph{varying} disturbances.}
        \label{fig:continuous_adapt_state}
    \end{subfigure}
    \hfill
    \hspace{0.1cm}
    \caption{Online adaptation of a state-based hovering policy to constant (a) and time-varying (b) disturbances. Vertical dashed lines indicate policy update steps which occur every 5\,\si{\second}. Shaded regions represent the error from the hovering target. In (b), each pair of solid vertical line and arrow indicates a direction reversal of the disturbance and the new direction.}
    \label{fig:sim_trajs}
    \vspace{-0.5cm}
\end{figure*}

For the state-based stabilizing hover task, we ran each method under all disturbance conditions from a set of 8 different starting positions around the hovering target, and used the final steady-state error as the performance metric. 
To ensure a fair comparison, we continued running each method until no further accuracy improvements were observed. This was found to be approximately 10 seconds for all baseline methods, as they do not require any policy adaptation, and around 30 seconds for our approach (both state and vision-based) for a few learning steps to take place.
As summarized in Tab.\,\ref{table:state_hovering_results}, results show that our method consistently exhibits superior or comparable performance to the baselines. 
Fig.\,\ref{fig:sim_hovering} illustrates that our method rapidly adapts the policy to compensate for the \textit{large} disturbances within 2-3 adaptation steps. 
DATT performs well under both the \textit{none} and \textit{small} disturbance scenarios, which are within its training distribution, but struggles to adapt to the larger, out-of-distribution disturbance. 

For visual feature-based hovering, the baseline methods would require an additional state estimation module, which limits the ability to clearly benchmark their performance without confounding errors from the state estimator, and are thus excluded from the comparison.
As shown in Tab.\,\ref{table:vision_hovering_results}, our visual feature-based approach resulted in larger errors than our state-based approach. 
We empirically observed that adaptation of the visual feature-based policy is less stable than the state-based counterpart and may require more policy learning epochs or update steps, most likely due to partial state observability and sample inefficiency in learning vision-based control. 
More detailed experiments and performance results are provided in the supplementary material.
\begin{table}[t]
\centering
\caption{Average tracking errors (in \si{\meter}) for \rebuttal{three} different trajectories (\textit{Circle}, \textit{Figure-8}, and \rebuttal{\textit{5-Point Star}}). The errors of the two best-performing methods for each disturbance condition are highlighted in green and orange.}
\label{table:sim_tracking_results}
\setlength{\tabcolsep}{2pt}
\vspace{-0.1cm}
\begin{tabular}{ccccc}
\toprule
Trajectory & Method & No Dist. & Small Dist. & Large Dist. \\ \midrule
\multirow{5}{*}{\textit{Circle}}
& Base DiffSim   & 0.365 $\pm$ 0.124   & 0.571 $\pm$ 0.091      & 1.479 $\pm$ 0.213      \\ 
 & $\mathcal{L}_1$-MPC & \cellcolor{Orange!10}0.113 $\pm$ 0.027   & \cellcolor{Orange!10}0.096 $\pm$ 0.063      & 0.410 $\pm$ 0.155       \\
& DATT (PPO)   & \cellcolor{YellowGreen!25}0.058 $\pm$ 0.016   & \cellcolor{YellowGreen!25} 0.040 $\pm$ 0.024       & \textit{crash}             \\
& Ours   & 0.167 $\pm$ 0.048   & 0.135 $\pm$ 0.101      & \cellcolor{Orange!10}0.349 $\pm$ 0.175      \\ 
& Ours (LoRA)   & 0.129 $\pm$ 0.051   & 0.159 $\pm$ 0.043      & \cellcolor{YellowGreen!25}0.326 $\pm$ 0.061      \\ \midrule
\multirow{5}{*}{\textit{Figure-8}}
&Base DiffSim   & 0.313 $\pm$ 0.087    & 0.492 $\pm$ 0.155       & 1.363 $\pm$ 0.382 \\
&$\mathcal{L}_1$-MPC & 0.109 $\pm$ 0.063   & 0.121 $\pm$ 0.025      & 0.281 $\pm$ 0.097      \\
&DATT (PPO)  & 0.078 $\pm$ 0.037   & 0.082 $\pm$ 0.046      & \textit{crash}             \\
&Ours   & \cellcolor{YellowGreen!25} 0.068 $\pm$ 0.040    & \cellcolor{YellowGreen!25}0.045 $\pm$ 0.03       & \cellcolor{Orange!10}0.137 $\pm$ 0.098      \\
&Ours (LoRA)   & \cellcolor{Orange!10} 0.069 $\pm$ 0.047    & \cellcolor{Orange!10}0.059 $\pm$ 0.043       & \cellcolor{YellowGreen!25}0.110 $\pm$ 0.037      \\ \midrule

\multirow{5}{*}{\textit{\rebuttal{5-Point Star}}}
&\rebuttal{Base DiffSim}   & \rebuttal{0.453 $\pm$ 0.190}    & \rebuttal{0.467 $\pm$ 0.236}       & \rebuttal{0.844 $\pm$ 0.389} \\
&\rebuttal{$\mathcal{L}_1$-MPC} & \rebuttal{0.295 $\pm$ 0.086}   & \rebuttal{0.218 $\pm$ 0.111}      & \rebuttal{0.417 $\pm$ 0.086}      \\
&\rebuttal{DATT (PPO)}  & \cellcolor{YellowGreen!25} \rebuttal{0.087 $\pm$ 0.059}   & \cellcolor{YellowGreen!25} \rebuttal{0.102 $\pm$ 0.093}      & \rebuttal{\textit{crash}}             \\
&\rebuttal{Ours}   & \cellcolor{Orange!10} \rebuttal{0.126 $\pm$ 0.094}    & \rebuttal{0.133 $\pm$ 0.075}       & \cellcolor{YellowGreen!25} \rebuttal{0.211 $\pm$ 0.116}      \\
&\rebuttal{Ours (LoRA)}   & \rebuttal{0.129 $\pm$ 0.069}    & \cellcolor{Orange!10} \rebuttal{0.130 $\pm$ 0.099}       & \cellcolor{Orange!10} \rebuttal{0.231 $\pm$ 0.076}      \\ 

\bottomrule
\end{tabular}
\vspace{-0.5cm}
\end{table}

For trajectory tracking, we recorded 60-second rollouts and computed the average tracking error (\si{\meter}) within the last 10-second window as the performance metric.
As shown in Tab.\,\ref{table:sim_tracking_results}, our approach achieves comparable performance to the baselines across all trajectories and disturbance conditions. 
\rebuttal{Here, our method exhibits consistent responsiveness to both smooth and non-smooth references. In particular, for the \textit{5-Point Star}, our method achieves comparable tracking accuracy as DATT, which has demonstrated strong capabilities in tracking non-smooth, dynamically infeasible trajectories~\cite{huang2023datt}.}
We observed that our method is able to rapidly adapt and achieve much improved tracking accuracy after only 3-4 policy update steps.
For DATT, consistent with findings from state-based hovering, it fails to generalize to the out-of-distribution disturbances and results in crashes for all reference trajectories.
\begin{table}[b]
\centering
\vspace{-0.3cm}
\caption{Average steady-state error (in \si{\meter}) from the hovering target across 8 rollouts. The errors of the two best-performing methods for each disturbance condition are highlighted in green and orange.}
\vspace{-0.1cm}
\label{table:vision_hovering_results}
\setlength{\tabcolsep}{2.5pt}
\begin{tabular}{cccc}
\toprule
Method        & No Dist. & Small Dist. & Large Dist. \\ \midrule
Base DiffSim (Vision) & 0.133 $\pm$ 0.009   & 0.404 $\pm$ 0.039      & 1.383 $\pm$ 0.176      \\ 
Ours (State)  & \cellcolor{YellowGreen!25} 0.015 $\pm$ 0.001   & \cellcolor{YellowGreen!25} 0.008 $\pm$ 0.002      & \cellcolor{YellowGreen!25} 0.105 $\pm$ 0.007      \\
Ours (Vision) & \cellcolor{Orange!10} 0.082 $\pm$ 0.009   & \cellcolor{Orange!10}0.099 $\pm$ 0.021      & 0.207 $\pm$ 0.041      \\ 
Ours (Vision, LoRA) & 0.084 $\pm$ 0.014   & 0.111 $\pm$ 0.024      & \cellcolor{Orange!10}0.205 $\pm$ 0.048      \\ \bottomrule
\end{tabular}
\end{table}

\vspace{-0.4cm}
Finally, we observe that using low-rank policy adaptation with our approach achieved comparable performances to using full adaptation across all tasks and disturbance conditions.
This demonstrates that LoRA can indeed be effectively combined with the differentiable simulation framework to achieve both sample- and parameter-efficient policy adaptation, where the latter may be particularly advantageous for fine-tuning pretrained policy networks that are significantly larger than the MLP used in our experiments.
\subsubsection{Computational and Sample Efficiency Analysis}
We analyze and compare the sample and computational efficiency of our approach against DATT for state-based tasks. 
For our approach, policy pretraining uses 300 epochs across 100 environments, which is equivalent to 4.5 million simulation steps in total, and takes approximately \emph{15 seconds}.
Empirically, we observed that good-performing initial policies can in fact be obtained using fewer epochs and environments, thanks to the low-variance first-order policy gradients from differentiable simulation. 
For online adaptation, each residual dynamics learning step (100 iterations) takes approximately \emph{2 seconds}, and each policy adaptation step runs for 30 epochs across 10 environments (or 45k simulation steps) and takes about \emph{1.5 seconds}. Here, with only 3 adaptation steps, which is equivalent to \textbf{4.5 seconds} of policy training in wall time, we already observe significant performance improvements for both hovering and tracking.
In comparison, DATT trains the policy for 20 million simulation steps, which takes around \emph{2 hours}, and requires no further training at runtime.
For DATT, we also observed slower convergence to lower rewards when training with larger domain randomization, which is likely a result of the performance-generalization trade-off~\cite{zhou2022domain}, possibly exacerbating the high variance in the policy gradient estimates.
In summary, our approach enables more efficient compute usage by simplifying initial policy training without domain randomization or curricula, and by supporting sample- and compute-efficient online adaptation to out-of-distribution scenarios where domain randomization fails to generalize.

\subsubsection{Continuous Adaptation to Time-Varying Disturbances}
We demonstrate the ability of our method to continuously adapt policies to unknown time-varying disturbances, using the realistic quadrotor simulator~\cite{foehn2022agilicious}. 
Here, we show an example of continuously adapting a state-based hovering policy under uniform, time-varying acceleration disturbance given by $\pm [0.5, 0.5, 0.5]\,\unit{m/s^2}$, which reverses its direction every 15\,\si{\second}. 
As shown in Fig.\,\ref{fig:continuous_adapt_state}, our approach rapidly adapts the state-based policy within 2 adaptation steps to adjust for each disturbance change, with the policy behavior remaining stable throughout the entire process.
We observed that the continuous adaptation of a visual feature-based hovering policy was less stable than its state-based counterpart, consistent with our previous findings.
For continuous trajectory tracking adaptation experiments, we provide detailed visualizations in the supplementary material.
\vspace{-0.3cm}

\begin{figure}[t]
    \centering
    \begin{subfigure}{0.436\columnwidth}
        \centering
        \includegraphics[width=\linewidth]{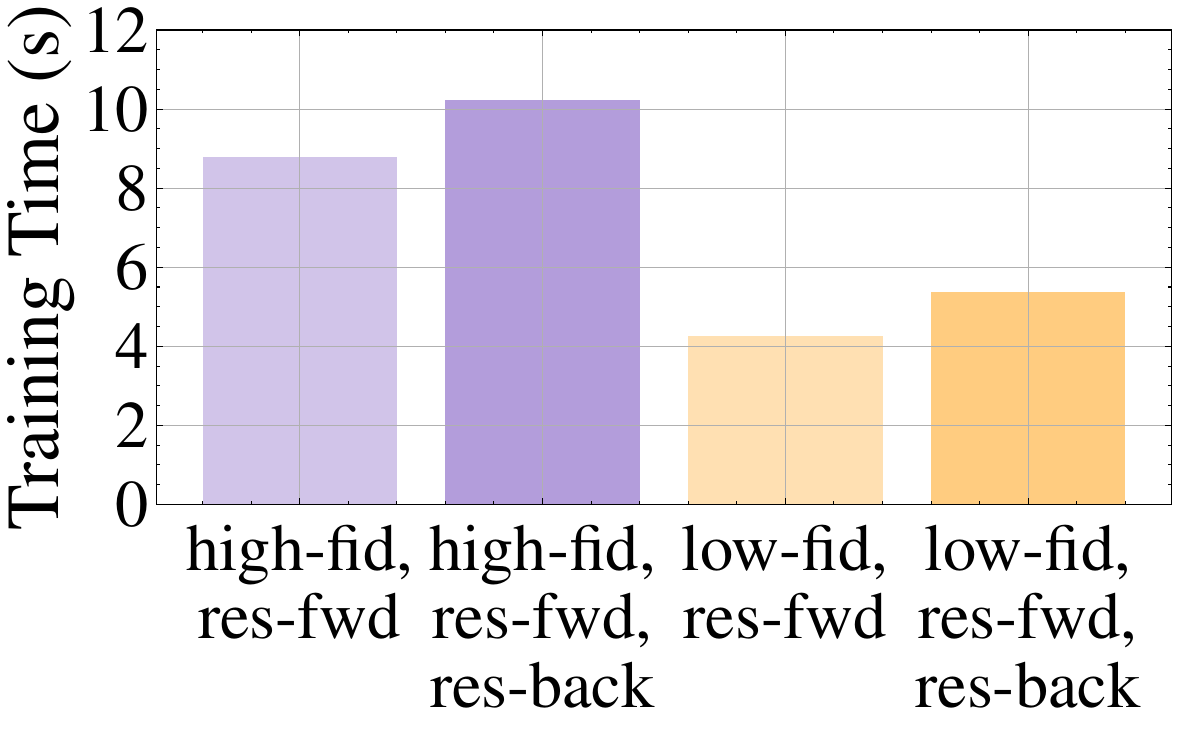}
        \vspace{-0.5cm}
        \caption{\footnotesize Training time}
        \label{fig:analysis_training_time}
    \end{subfigure}
    \hfill
    \begin{subfigure}{0.54\columnwidth}
        \centering
        \includegraphics[width=\linewidth]{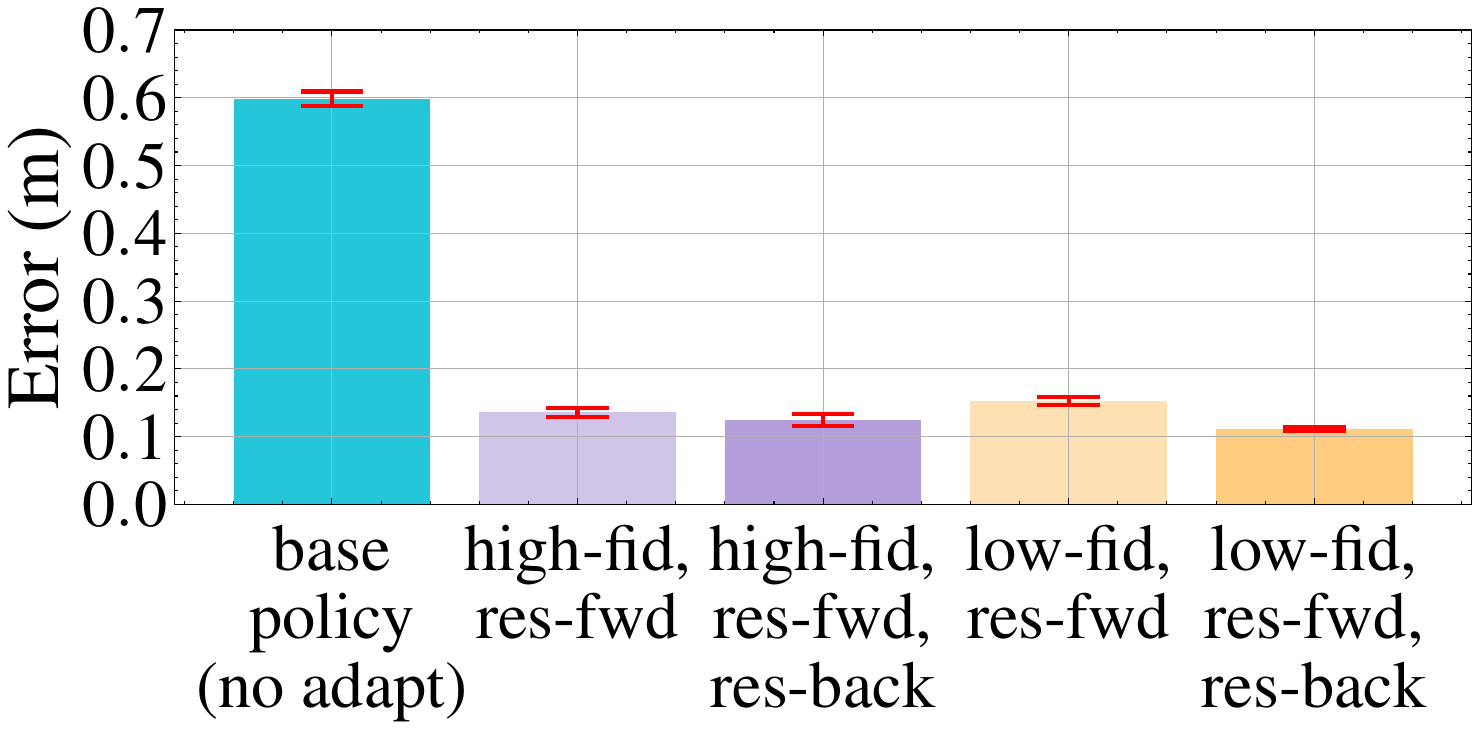}
        \vspace{-0.5cm}
        \caption{\footnotesize Final steady-state error}
        \label{fig:analysis_policy_performance}
    \end{subfigure}
    \vspace{-0.5cm}
    \caption{(left) Policy training times using four different simulation configurations. (right) Resulting policy performances compared against the base policy performance. Error bars show $\pm$3 standard deviations of the error distribution across 8 rollouts for each configuration.}
    \label{fig:analysis_results}
    \vspace{-0.6cm}
\end{figure}

\begin{figure*}[b]
    \centering
    \hfill
    \begin{subfigure}{0.64\columnwidth}
        \centering
        \includegraphics[width=\linewidth]{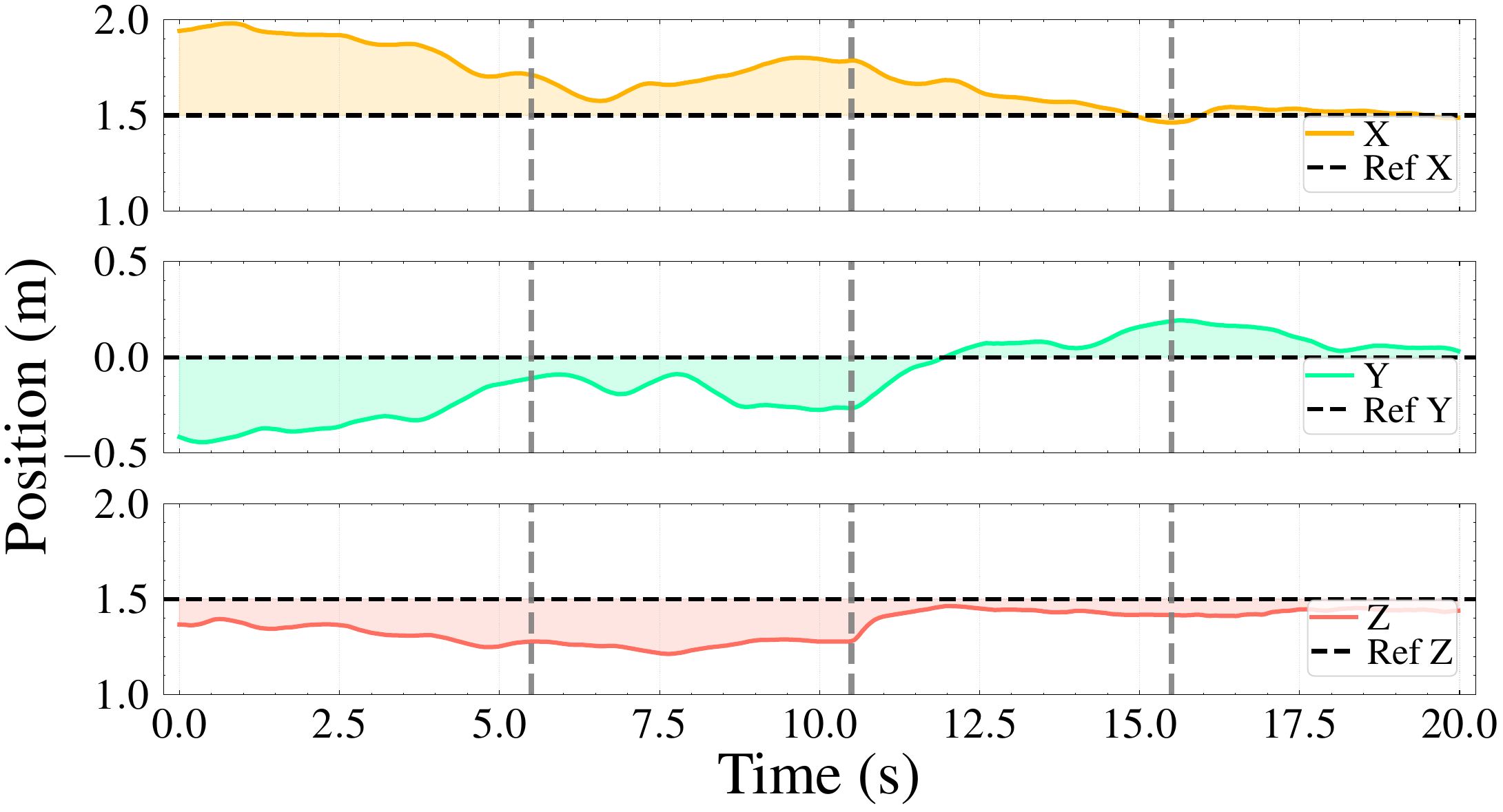}
        \vspace{-0.5cm}
        \caption{\footnotesize State-based hover with added mass and wind.}
        \label{fig:real_hovering_traj}
    \end{subfigure}
    \hfill
    \begin{subfigure}{0.64\columnwidth}
        \centering
        \includegraphics[width=\linewidth]{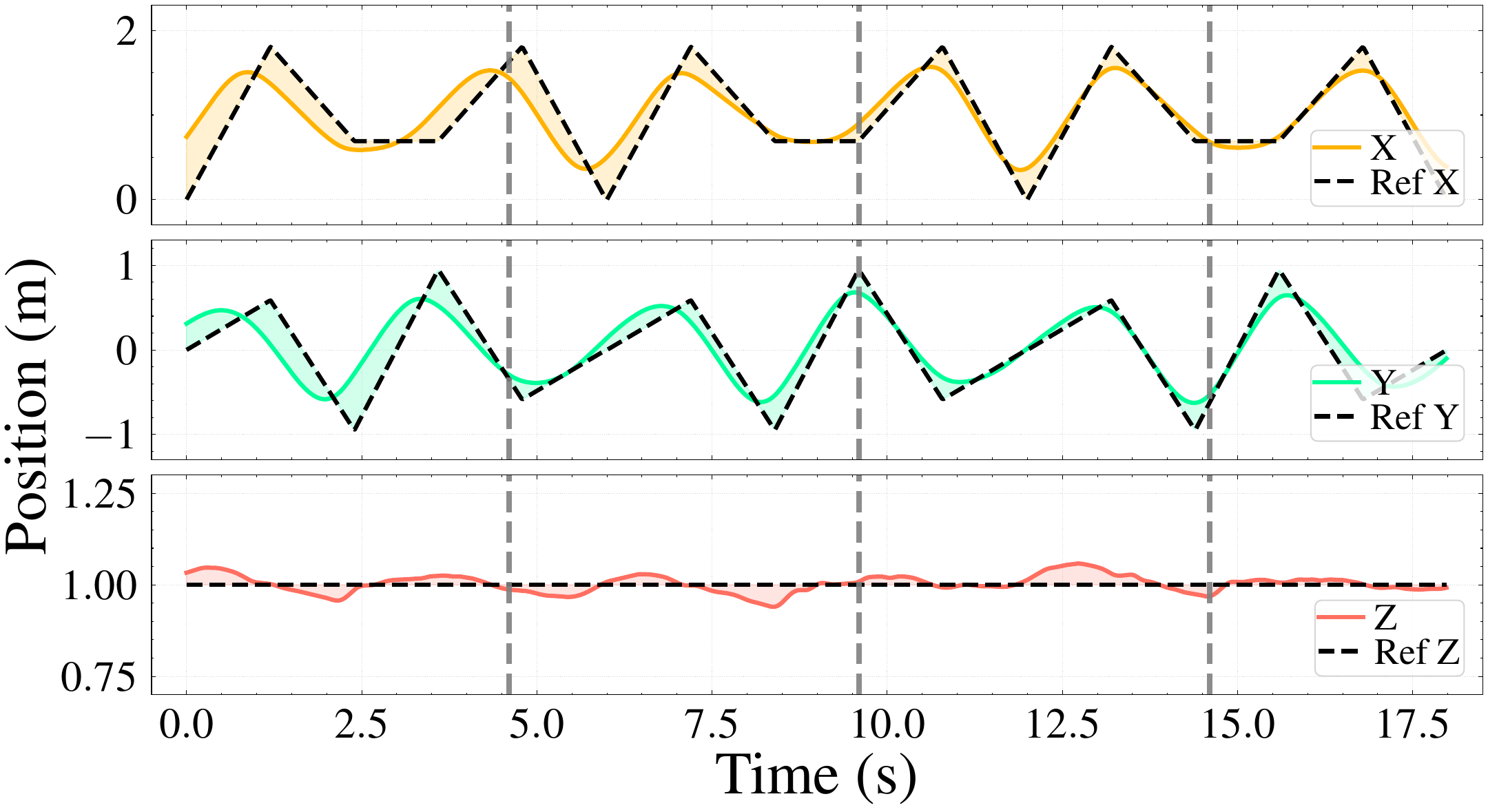}
        \vspace{-0.5cm}
        \caption{\footnotesize \rebuttal{\textit{5-Point Star} tracking with added wind.}}
        \label{fig:real_star_traj}
    \end{subfigure}
    \hfill
    \begin{subfigure}{0.64\columnwidth}
        \centering
        \includegraphics[width=\linewidth]{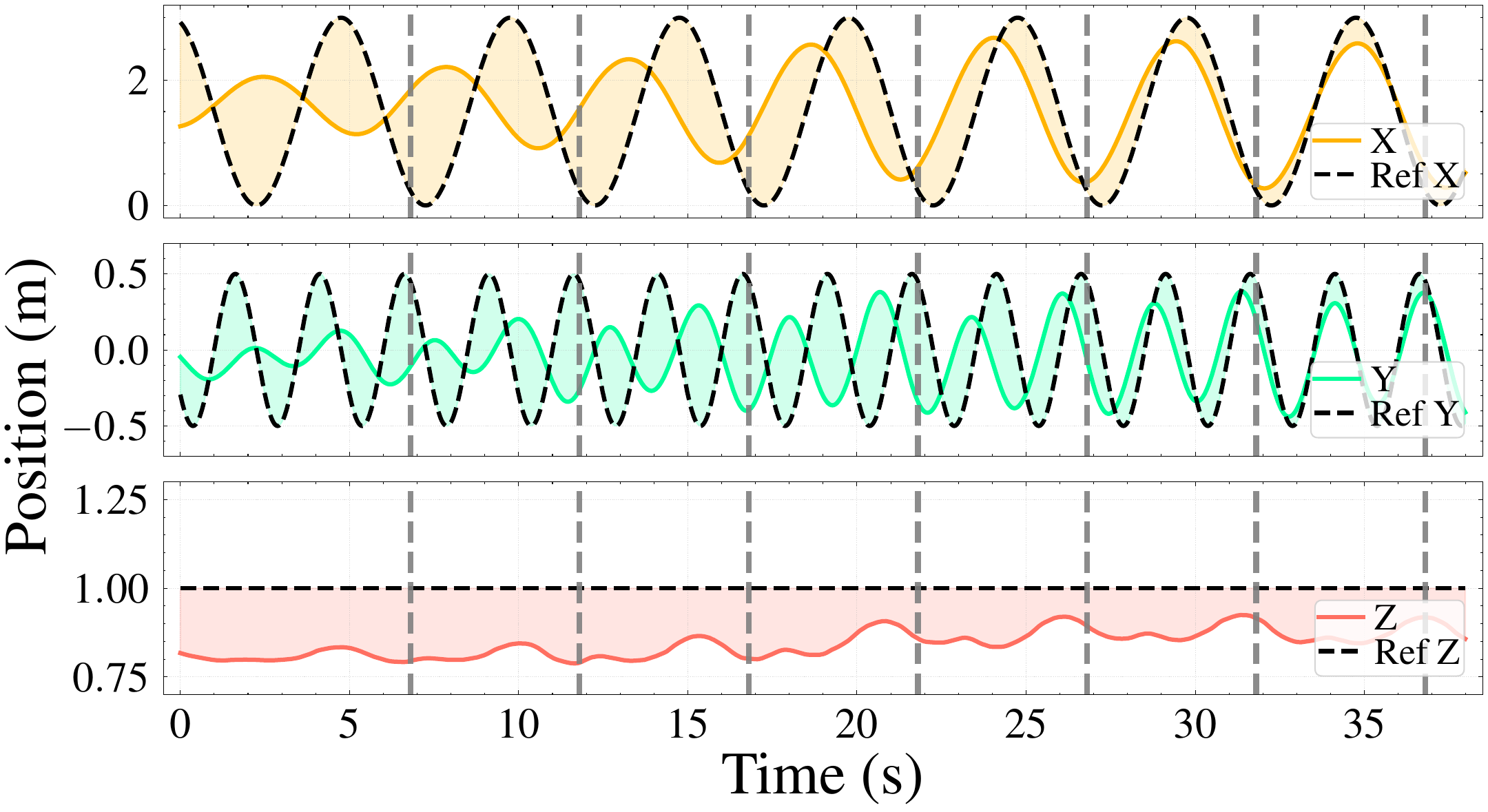}
        \vspace{-0.5cm}
        \caption{\footnotesize \textit{Figure-8} tracking on the large quadrotor.}
        \label{fig:real_fig8_traj}
    \end{subfigure}
    \hfill
    \vspace{-0.1cm}
    \caption{Rapid policy adaptation using our proposed approach in the real world. Vertical dashed lines indicate policy update steps which occur every 5\,\si{\second}. Shaded regions represent the error from the reference.}
    \label{fig:real_experiment_trajs}   
    \vspace{-0.3cm}
\end{figure*}
\subsection{Optimizing Runtime Efficiency and Performance}
\label{subsec:design_choice_justification}
We conducted an analysis using the state-based stabilizing hover task to justify two key design choices in the differentiable simulation pipeline: 1) low-fidelity analytical dynamics model for simulation, and 2) gradient backpropagation only through the analytical dynamics model. 
Given that a key objective is to minimize runtime while maintaining policy performance, we compared the training time and policy performance for each design choice. We used the same realistic quadrotor simulator~\cite{foehn2022agilicious} and added a constant uniform acceleration disturbance of 2\,$\unit{m/s^2}$ in the positive x-axis direction. We first collected 50 3-second rollout trajectories of the base hovering policy from random starting positions around the target, and then used the generated residual samples to train a single residual dynamics network for 200 epochs. Finally, we adapted the base policy by running 100 epochs of policy training across 100 parallel environments, and evaluated the final steady-state errors from the hovering target across 8 rollout trajectories.

We first compared using the low-fidelity (low-fid) dynamics model \eqref{eqn:simple_quad_dynamics} against using a high-fidelity model (high-fid), which simulates body and rotor drag effects, rotor thrust maps, and low-level controller dynamics, as the analytical dynamics model together with the residual dynamics network for forward simulation (res-fwd). We found that using the low-fidelity model achieves approximately 2-times faster training (see Fig.\,\ref{fig:analysis_training_time}) than using the high-fidelity model, while the achieved policy performances by both methods were very comparable (see Fig.\,\ref{fig:analysis_policy_performance}). For gradient backpropagation, we found that in addition to backpropagating through the analytical dynamics model, also performing backpropagation through the learned residual dynamics network (res-back) increases training time by approximately 30\% without providing clear benefits to the policy performance. This is consistent with previous findings~\cite{heeg2024learning,song2024learning} that combining accurate forward simulation with a surrogate gradient that points in approximately the same direction as the true gradient vector accelerates policy training without impacting the resulting policy performance.

\begin{table}[t] 
\centering 
% \vspace{-0.3cm}
\caption{Quadrotor parameters for simulation and real-world experiments.} 
\vspace{-0.1cm}
\label{table:drone_params}
\setlength{\tabcolsep}{4pt}
\fontsize{8pt}{10pt}\selectfont
\begin{tabular}{ l  c c}
    \toprule
      \textbf{Param.}  & \textbf{Small Quadrotor}  & \textbf{Large Quadrotor}\\ \cline{1-2}
      \grayrow
      Mass [\SI{}{\kilogram}]  &0.19 &    0.60\\
      Maximum Thrust [\SI{}{\newton}] &14.00 & 34.00   \\
      \grayrow
      Arm Length [\SI{}{\meter}] &0.06 & 0.13 \\
      % \grayrow
      Inertia  [\SI{}{\gram\square\meter}] &[0.14, 0.17, 0.21] & [2.41, 1.80, 3.76] \\
      \grayrow
      Motor Time Constant [\SI{}{\second}] &0.025 &0.033 \\
    \bottomrule
\end{tabular}
\vspace{-0.5cm}
\end{table}

%%%%%%%%%%%%%%%%%%%%%%%%%%%%%%%%%%%%%%%%%%%%%%%%%%%%%%%%%%%%%%%%%%
\vspace{-0.4cm}
\subsection{Real World Validation}
We conducted real-world experiments using the same tasks as the simulated experiments and the same pretraining and online adaptation procedures.
A motion-capture system provides quadrotor state estimation at 100\,\si{\hertz} to the off-board workstation, which computes and sends commands to the on-board controller at 50\,\si{\hertz}. 
We used two quadrotors adapted from the Agilicious platform~\cite{foehn2022agilicious}: a small, lightweight quadrotor and a larger, heavier one with different dynamical properties (see Tab.\,\ref{table:drone_params}).
Moreover, we modified the small quadrotor by rigidly attaching to it a quadrotor stand from below, increasing its mass from 190\,\si{\gram} to 260\,\si{\gram} by approximately $37\%$ and altering its inertial properties. 
Finally, we used a fan to create wind disturbances, resulting in complex, state-dependent forces on the modified quadrotor due to its highly imbalanced and non-uniform drag profile.
Both the existing sim-to-real gap and the extra disturbances contribute to significant out-of-distribution dynamics that were unseen during policy pretraining.

Fig.\,\ref{fig:real_hovering_traj} shows state-based hovering adaptation on the modified small quadrotor under a diagonal wind disturbance. 
Despite the more complex and unstable real-world disturbance forces compared to the constant uniform disturbance in simulation, our method still enables the policy to rapidly adapt with 2-3 policy update steps to compensate for disturbances.
Similar results were observed for visual feature-based hovering, where the adaptation process appeared less stable than state-based hovering, which is consistent with our findings from simulated experiments.
Real-world experiments also show that our approach achieves accurate trajectory tracking under \rebuttal{added mass, wind, and} significant model mismatches. 
\rebuttal{In particular, our method achieves accurate tracking of the non-smooth \textit{5-Point Star} under complex wind disturbances (see Fig.\,\ref{fig:real_star_traj}).}
Moreover, Fig.\,\ref{fig:real_fig8_traj} shows one particular experiment where a \textit{Figure-8}-tracking policy was deployed on the large quadrotor. 
Despite the poor initial tracking and state-space exploration caused by the large sim-to-real gap, the policy quickly adapts and achieves much improved tracking within just a few policy update steps. 
\rebuttal{Videos and visualizations of real-world experiments are provided in the supplementary material.}
\vspace{-0.1cm}

\section{Conclusion}\label{sec:conclusion}
We propose a novel rapid policy adaptation framework combining online residual dynamics learning from real-world flight data and sample-efficient policy learning via differentiable simulation.
With all system components designed for rapid adaptation, we demonstrate the possibility to adapt both state and visual feature-based policies to unknown disturbances within \emph{several seconds}. 
One limitation of our framework lies in the tightly-coupled dependencies between data collection via policy rollout and policy learning using learned residual dynamics from the collected data.
The quality and rate of convergence may be affected by biases or noise in the learned residual dynamics.
\ifthenelse{\equal{\isArXiv}{0}}{
The dependency is closely related to the concept of \emph{performative prediction}~\cite{perdomo2020performative} in related machine learning fields.
}{}
Thus, future work will explore uncertainty-driven data collection where the policy is augmented by active exploration to simultaneously improve task performance and reduce uncertainty in the real-world dynamics. 
%

%
% \addtolength{\textheight}{-12cm}   % This command serves to balance the column lengths
                                  % on the last page of the document manually. It shortens
                                  % the textheight of the last page by a suitable amount.
                                  % This command does not take effect until the next page
                                  % so it should come on the page before the last. Make
                                  % sure that you do not shorten the textheight too much.

%
% \input{chapters/appendix}
%
% \section*{ACKNOWLEDGMENT}
\balance
\bibliographystyle{ieeeTran}
\bibliography{ral}

% \clearpage
%
% \input{chapters/appendix}

\end{document}